\documentclass[cameraready]{Interspeech}
\title{Do speech foundation models really learn words?}

\author[affiliation={1}]{Robin}{Huo}
\author[affiliation={1}]{Ewan}{Dunbar}
\address{ $^1$ University of Toronto, Canada }

\email{robin.huo@mail.utoronto.ca, ewan.dunbar@utoronto.ca}


\usepackage{comment}

\begin{document}

\maketitle

\begin{abstract}
    Self-supervised speech foundation models are now used in a wide array of downstream applications, including traditional speech recognition and as the basis for tokens in speech-aware language models. Attempts to understand their usefulness have largely focused on probing their representations' ability to discriminate phonemes and words. However, discriminative ability for words need not imply specialized representation of words per se. Good discrimination of words may be explained by good encoding of word form (phonemes) rather than form-independent word representations encoding identity or syntactic/semantic properties. By partialling out phoneme information using residualization, we show that, in later layers, HuBERT and wav2vec 2.0 do in general learn representations which encode words with reasonable fidelity independently of local phonetic content. We show that this simple approach to disentanglement can enhance higher-order linguistic information in word discovery tasks.
\end{abstract}

\section{Introduction}

Speech foundation models based on self-supervised learning (S3M) now form the foundation of approaches for various speech-related downstream tasks \cite{mohamed_self-supervised_2022}. The success of S3Ms, similar to the success of large language models, is generally understood to imply that success on the prediction task demands that the learner construct abstract representations  corresponding to a latent structure for language, in much the same way that simple contextual word embeddings discover lexical semantics. As such, a number of probing studies have been interpreted to mean that these models learn representations that encode phonemes, speakers, and words \cite{martin_probing_2023, pasad_what_2024, shen_wave_2023, english_domain-informed_2022, van_niekerk_analyzing_2021}.

However, success in probing ``words,'' in the sense of mere word identity, is difficult to interpret. A token of a word like \emph{cat} is both an instance of a bundle of syntactic and semantic features (a noun referring to a fuzzy four-legged animal) as well as an instance of the sequence of phonemes /kæt/. In Saussure's \cite{saussure1916course} terms, it is an instance of both the signified and the signifier. Since it seems clear that trained S3M representations are highly discriminative of phonemes, successful probes of word identity \cite{pasad_comparative_2023} could be explained away by models' ability to discriminate phonemes. Although some work has found correlation with text-based word vectors \cite{pasad_layer-wise_2021}, as well as part-of-speech and semantic tags  \cite{pasad_what_2024}, the ability to predict this richer lexical information is not a decisive sign that models learn anything beyond than phonological form: a representation encoding phonological form allows words to be identified (mostly) uniquely, and \emph{properties} of  words can be predicted from  word identity.

 Choi et al.\ \cite{choi_self-supervised_2024} attempted to address this by comparing near-homophones with synonyms, with synonyms showing much weaker cosine similarities, suggesting partial but generally weak encoding of non-phonological (``signified'') properties of words. This result suffers from an issue common to studies using cosine similarities, and, more generally, any similarity calculated indiscriminately over  representational dimensions: information can be ``drowned out'' by irrelevant dimensions, even if it is represented with high fidelity on a subspace. Nevertheless, the result is consistent with the fact that S3Ms lag behind text-based models in semantic-based tasks \cite{ashihara_unveiling_2024}.

 This study takes a simple residualization approach to examining the question of whether English-pretrained HuBERT and wav2vec 2.0 models encode  non-phonological lexical information. We use a linear classifier, rather than a global similarity-based approach, predicting word identity at the frame level without pooling, similar to Pasad et al.\ \cite{pasad_what_2024}. We then remove a component of the representation that encodes phonological information. We demonstrate that, while much of the ability to identify words indeed disappears from the resulting residualized representations, in most cases, later layers are still predictive of word identity. We conclude that information is integrated from the surrounding context, essentially creating a ``token'' that allows for words to be identified locally at any frame throughout the word. While this does not tell us whether and how much lexical semantic or syntactic information is encoded, we demonstrate that the representations encode word information that goes beyond encoding diphones or triphones.

\section{Related methods}

We take a simple residualization  approach to ``subtracting'' information from S3M representations for the purposes of interpretation. This approach involves the estimation of a linear function predicting the embedding features given target labels, and subtraction of this function's predicted embeddings from the true embeddings to yield residualized embeddings where information about the targets has been linearly removed. In our case, the function predicts embedding features given categorical phoneme labels, and thus the residualization step can roughly be seen as subtracting phoneme-level means. This intervention on model features enables direct experimental investigation of the contribution of a given kind of information to model behaviour on downstream tasks or probes that are fundamentally linear.

Toneva et al.\ \cite{toneva_combining_2022} used a similar method to remove context-independent word token contributions to contextual ELMo embeddings  and evaluated the resulting residualized text embeddings against alignment with brain imaging data; Oota et al.\ \cite{oota_joint_2023} performed a similar analysis with BERT \cite{devlin_bert_2019}. Oota et al.\ \cite{oota_speech_2024} evaluated a wider variety of text models and extended this analysis to speech models wav2vec~2.0 \cite{baevski_wav2vec_2020} and Whisper \cite{radford_robust_2022}.

This approach can in principle lead to errors in both directions: failure to fully remove phoneme information when it is present, and removal of more than just phoneme information. Our primary concern is the first, as, for the purposes of our experiment, removing ``too much'' information would only strengthen our conclusion. We discuss how we address these issues in Section \ref{sec:method}, and demonstrate in Section \ref{sec:results} that the method is largely effective in removing phoneme information.

Two alternatives stand out. Many disentanglement methods work by building parallel representations which are expressly designed to modularly specialize in predicting a specific kind of information such as rhythm \cite{chan2022speechsplit2}. This is not appropriate for our goal, which is to understand and make use of existing representations which were not in general trained in such a way.

The approach taken by \cite{liu_self-supervised_2023} to remove known factors from S3M representations for the purposes of interpretation is to identify ``subspaces'' which appear to encode (for example) speaker information, and then collapse the corresponding axes out of the representations. However, the axes found are not true subspaces \textit{per se}. Instead, the method finds axes and associates an ``importance'' score to each, ostensibly measuring the amount of information about the factor encoded on the given dimension. There are two problems with this approach. First, it requires selecting some arbitrary number of dimensions to remove that is, in general, strictly less than the representation dimension. Second, it requires that we select an appropriate importance score. While \cite{liu_self-supervised_2023} used explained variance across factor levels, variance is not a full substitute for information, and thus does not guarantee that we have removed factor information. A variety of alternative methods exist for assigning importance to a dimension with respect to some factor (e.g., \cite{geyer_measuring_2024}), but the question of which to use in this context, if any, has not been explored.

\section{Method}
\label{sec:method}

We analyze representations from the final convolutional layer and each of the 12 transformer layers of the base variants of the SSL speech models HuBERT \cite{hsu_hubert_2021} and wav2vec~2.0 \cite{baevski_wav2vec_2020}, using the pre-trained checkpoints available from the fairseq repository.\footnote{\url{https://github.com/facebookresearch/fairseq/blob/main/examples/hubert/README.md}} We evaluate on the dev-clean split of LibriSpeech (English) \cite{panayotov_librispeech_2015}.  Following \cite{pasad_comparative_2023,pasad_layer-wise_2021}, we use alignments to associate each frame with a gold phoneme (42 categories, 3 being nonspeech/silence) and word label (8217 words + 1 silence category), and probe these frames using a softmax classifier. Before residualization and probing, we scale and center model representations to zero mean and unit variance. Past work suggests this transformation does not harm encoding of phoneme or word information, and may in fact enhance it \cite{nakamura_discrete_2025}. Although these representations are standardized, we refer to them henceforth as ``raw'', to contrast them with the residualized representations.

We measure word identity classification accuracy and assess the effect of residualizing out a factor corresponding to phonemic information. We test four variants: phonemes; diphones, both left- and right-hand; and triphones. That is, for a given frame, we residualize out the phoneme (or diphone/triphone) corresponding to that frame, and use the resulting representation as the input to the word probe. 
To perform the residualization, we estimate a ridge regression predicting individual representation frames from a one-hot encoding of the factor (phoneme, left-diphone, right-diphone, or triphone). We select the weight decay coefficient $\alpha$ by searching over values logarithmically ranging between $1$ and $0.0001$. Our goal is effectively to estimate the mean for each factor level, with a small amount of regularization.  We then subtract the predictions of the regression. In estimating the regression models, we exclude any examples of phonemes, diphones, or triphones which correspond exactly to, or properly contain, the complete word corresponding to that frame.
Frames labelled with phoneme categories corresponding to non-speech (SIL, SP, SPN) are retained when estimating the residualization models and discarded when analyzing results.

We train and evaluate a linear probe predicting word identity with frames as input using five-fold cross-validation. To validate the method and ensure that phoneme information is not recoverable linearly after the residualization procedure, we evaluate a linear probe predicting phoneme identity.

\begin{table*}[ht]
    \caption{Accuracy of a linear probe trained to predict phonemes from HuBERT and wav2vec~2.0 representations: raw, residualized from phoneme labels as in the paper, and residualized without standardization.}
    \centering

\begin{tabular}{llccccccccccccc}
    \toprule
    \textbf{Model} & \textbf{Features} & \textbf{Conv} & \textbf{1} & \textbf{2} & \textbf{3} & \textbf{4} & \textbf{5} & \textbf{6} & \textbf{7} & \textbf{8} & \textbf{9} & \textbf{10} & \textbf{11} & \textbf{12} \\
    \midrule
    HuBERT & Raw & 0.46 & 0.70 & 0.72 & 0.75 & 0.77 & 0.82 & 0.84 & 0.85 & 0.86 & 0.86 & 0.87 & 0.87 & 0.87 \\
    & Residual & 0.78 & 0.48 & 0.34 & 0.29 & 0.29 & 0.29 & 0.29 & 0.28 & 0.27 & 0.28 & 0.27 & 0.29 & 0.36 \\
 & Res. w/o std. & 1.00 & 0.97 & 0.90 & 0.77 & 0.75 & 0.80 & 0.71 & 0.76 & 0.70 & 0.58 & 0.55 & 0.68 & 0.82 \\
    \midrule
wav2vec~2.0 & Raw & 0.51 & 0.72 & 0.75 & 0.78 & 0.81 & 0.83 & 0.84 & 0.84 & 0.84 & 0.85 & 0.84 & 0.74 & 0.74 \\
 & Residual & 0.35 & 0.32 & 0.29 & 0.28 & 0.28 & 0.27 & 0.26 & 0.27 & 0.27 & 0.28 & 0.29 & 0.60 & 0.51 \\
 & Res. w/o std. & 1.00 & 0.89 & 0.72 & 0.68 & 0.71 & 0.67 & 0.57 & 0.58 & 0.53 & 0.66 & 0.77 & 0.98 & 0.99 \\
    \bottomrule
\end{tabular}
    \label{tab:results-val}
\end{table*}

\begin{figure*}[ht]
    \centering
    \includegraphics[width=1.01\textwidth]{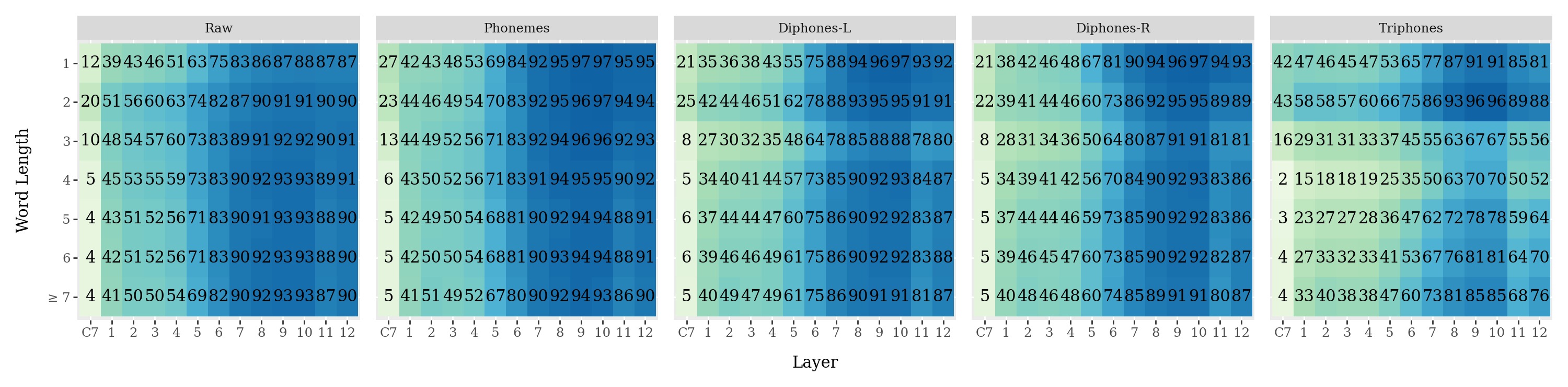}
    \includegraphics[width=1.01\linewidth]{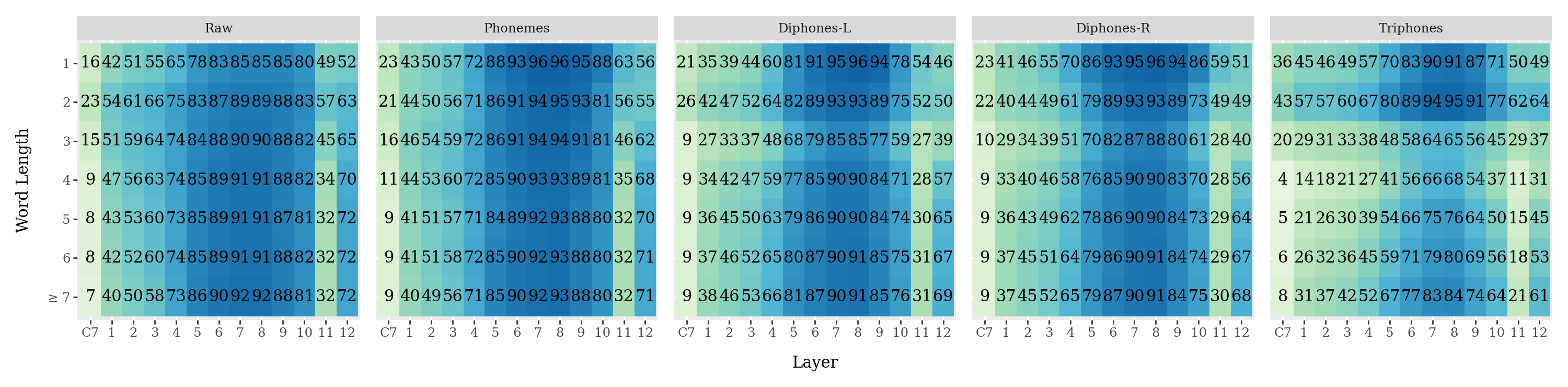}
    \caption{Accuracy (percent) of a linear probe predicting word identity from HuBERT (upper) and wav2vec~2.0 (lower) representations.}
    \label{fig:results}
\end{figure*}

\section{Results}
\label{sec:results}

Table \ref{tab:results-val} shows the results of the validation experiment (probe predicts phoneme rather than word identity). Results are accuracies averaged over five cross-validation folds. As can be seen, the residualization method is successful in making it substantially more difficult for the linear probe to classify individual phonemes. (The most common phoneme label has a relative frequency of 11.6\%; we discuss the fact that the method falls short of reaching chance level in Section \ref{sec:limitations} below.) We also include an ablation without the standardization step. The results show that this step is necessary, presumably because the shape of the distribution is not matched across phonemes, which can be exploited for classification. In the main experiment we always include the standardization step.

Figure \ref{fig:results} shows the results of the main experiment, divided by layer and by word length, with HuBERT in the upper panel and wav2vec 2.0 below. We observe that residualizing out phonemes   does not change the overall pattern seen in the raw representations: little ability to classify words in the last convolutional layer, and a sharp increase in the transformer layers peaking in layers 9 and 10 (HuBERT) or layers 7 and 8 (wav2vec 2.0) at somewhat more than 90\% accuracy. Removing diphones decreases accuracy slightly  more in earlier layers, across both models. 

Removing triphones decreases accuracy substantially except in layers 9 and 10 (HuBERT) and 7 and 8 (wav2vec 2.0), where it shows a mixed picture. While there is a substantial drop in the ability to classify words of lengths 3--6, the performance is still well above the baseline obtained by picking the modal word by length (length 3: 14.9\%; 4: 0.4\%; 5: 0.1\%; 6: $\ll$ 0.1\%) or by triphone (50.3\%). Thus, in some layers, the probe's ability to classify words is not merely a reflection of an encoding of short sequences of phonemes: the model encodes words partly independently of local phonological information.\footnote{At least for diphones and triphones, the fact that we did not include examples that contain full words when fitting the ridge regression is important here.  When we include them, performance drops substantially on words of length $\le$ 2 and 3 respectively. This is because in some cases the regression is fit to ``triphones'' which are \emph{also}  words.}
It is also worth noting that although the residualization does not completely reduce phonemic information to chance level as measured by a linear probe, the fact that Figure \ref{fig:results} shows significant differences where Table \ref{tab:results-val} shows largely uniform performance in the medial layers strongly suggests that the phoneme probe and word probe are leveraging different information.

\section{Word discovery}

\begin{table}
    \caption{Word discovery performance with HuBERT-base representations from layer 9, before and after residualization.}
    \centering
%    \begin{tabular}{lcccccc}
    \begin{tabular}{llccc}
        \toprule
        Hypers & Features & NED~$\downarrow$ & $F_1\uparrow$ & $R\uparrow$ \\
        \midrule
        % window size = 4, prom = 0.75
        Malan et al.\ \cite{malan_unsupervised_2025} & Raw & .508 & .162 & \textbf{.513} \\
            & $-$ Phoneme & \textbf{.463} & \textbf{.169} & \textbf{.513} \\
            & $-$ Diphone-L & .490 & .162 & .506 \\
            & $-$ Diphone-R & .474 & .160 & .502 \\
            & $-$ Triphone & .556 & .136 & .474 \\
        \midrule
        % window size = 4, prom = 0.15
        Best NED (for Raw) & Raw & .443 & .175 & .249 \\
            & $-$ Phoneme & \textbf{.439} & \textbf{.178} & \textbf{.258} \\
            & $-$ Diphone-L & .475 & .166 & .241 \\
            & $-$ Diphone-R & .474 & .169 & .236 \\
            & $-$ Triphone & .561 & .138 & .159 \\
        \midrule
        % window size = 5, prom = 0.55
        Best $R$ (for Raw) & Raw & .527 & .166 & .531 \\
            & $-$ Phoneme & \textbf{.478} & \textbf{.175} & \textbf{.537} \\
            & $-$ Diphone-L & .500 & .161 & .522 \\
            & $-$ Diphone-R & .505 & .155 & .511 \\
            & $-$ Triphone & .572 & .124 & .467 \\
         \bottomrule
    \end{tabular}
    \label{tab:results-lexicon}
\end{table}

To test whether the linear removal of phonemic information from S3M representations can benefit tasks which rely on encoding of higher-level linguistic information, we residualize phoneme target information out of the hidden representations and evaluate the performance on unsupervised word segmentation and clustering. We use HuBERT since its performance was better in the previous experiment.
We follow the method of \cite{malan_unsupervised_2025}: a word boundary detection algorithm \cite{pasad_what_2024} followed by k-means clustering into hypothesized word categories.

We perform this evaluation on LibriSpeech dev-clean and produce residualized phoneme representations using labels from the same alignments used in Section \ref{sec:method}, for the ninth layer of HuBERT-base.
We use this layer because it was found by \cite{pasad_comparative_2023} to display the strongest alignment with word information, and it is commonly used when training higher iterations on targets from HuBERT-base embeddings \cite{hsu_hubert_2021, chen_wavlm_2022}.
Following \cite{malan_unsupervised_2025}, we fix $k=13967$ in the clustering step for evaluation on LibriSpeech.
The boundary detection step algorithm uses two hyperparameters, a window size and a prominence threshold for peak detection over dissimilarity between adjacent S3M frames, for which we perform grid search (window size: 3--8 in increments of 1; prominence threshold 0--1 in increments of 0.05).

The results are evaluated according to the true word boundaries and types defined by the alignments, with a 20~ms tolerance for boundary placement.
Boundary quality is measured by token $F_1$-score, which describes how many of the true word tokens were correctly segmented (bounded correctly on both sides), and $R$-value, which captures the distance from the ideal operating point of perfect recall on boundary placement (hit-rate) and zero oversegmentation \cite{rasanen_improved_2009}.
The quality of the discovered lexicon is measured by normalized edit distance (NED), which is the average Levenshtein distance between the phonemic transcriptions (according to the gold alignments) of each token pair within a discovered word class, normalized by the length of the longest transcription.

Table \ref{tab:results-lexicon} shows the results across several hyperparameter settings.
Residualization from phoneme targets improves NED, token $F_1$ and $R$-value, reflecting more accurate word segmentation and clustering after the removal of phoneme information.
This improvement is robust across segmentation hyperparameter settings, which is important in particular for the task of unsupervised lexicon discovery as it aims to address low- or zero-resource circumstances in which there may be very little language data available to facilitate hyperparameter tuning.
It should be noted, however, that to directly use our approach requires aligned phonemic labels, which are not typically available in practice.
These results should thus be viewed not as an indication of practical performance, but as a validation of the core insight that direct removal of phonemic information may enhance the accessibility of other information.

Residualization from diphones and triphones, unlike residualization from phonemes, degrades segmentation performance.
Since phoneme transition probabilities are predictive of word boundaries \cite{saffran_1996_statistical}, it is likely that any encoding of local phonemic context aids segmentation irrespective of non-formal encoding of words themselves, and the removal of such information can thus harm segmentation performance as observed here.

\section{Discussion}

This experiment demonstrates that, in later layers, HuBERT and wav2vec~2.0 construct representations (that are predictive) of words and which are fairly robust to the erasure of information about phonemes or short sequences of phonemes. While it does not tell us what the model encodes about words, it strongly suggests that it treats them as units. This is consistent with the results of \cite{pasad_what_2024} that several S3Ms can be used to reconstruct word boundaries and, to some degree, higher-order properties of words. The novelty of our study is to remove the confound caused by the fact that the model also encodes local phonemic content and context, which is likely to be strongly predictive of word identity due to the Saussurean duality of words.

We limited our residualization to phoneme sequences of length $\le$ 3 in order to avoid the unreliable regression model fits that would result from moving to higher-order $n$-grams. It is  both  possible and plausible that the model integrates context from much longer distances---enough for a linear probe to predict  word identity---but still does not construct context-independent representations of specific words per se. Given that HuBERT is trained to predict masks spanning at least 10 frames (200~ms), which includes many full words, the fact that enough context is integrated to identify a word might seem like a trivial property that follows from the loss. However, given an average phoneme duration of around 65~ms, a single masked span contains only around 3 phonemes. Furthermore, the prediction targets are derived from clustering individual, contextless frames. It would have been entirely possible, therefore, for the model to construct representations that are predictive of local phonemic context only. In spite of this, they predict words, even quite long ones, with fairly high accuracy.

Our simple residualization approach largely succeeds in removing representations' ability to predict individual phonemes while maintaining their ability to predict word identity. In many situations, this is likely to be useful for enhancing higher-level information and reducing the influence of local phonological information. Clearly, as Figure \ref{fig:results} demonstrates, linear probing is not one of those situations: the raw representations already perform well on the word identity probe. However, as we show, lexicon discovery does benefit from phoneme residualization. This demonstrates the importance of complementing probing tasks with more complex downstream tasks.

\section{Summary of contributions}

We have shown that HuBERT and wav2vec~2.0, pre-trained on English, encode substantial information about word identity (or sufficient to reconstruct word identity) locally, within the frames aligned with a given word. While previous studies have reached similar conclusions, our study clearly dissociates the encoding of words from the encoding of phonological form:  encoding of local phonological features does not fully explain these S3Ms' ability to distinguish words. The result underscores the need to understand why self-supervised learning is so effective in uncovering word information from speech,  and what the limitations of these representations are, in particular in light of the serious gap between text and discrete speech units derived from S3Ms in unsupervised tasks such as lexicon discovery  \cite{malan_should_2026} and textless language modelling \cite{poli_spidr_2025}. A straightforward application of our result improves performance in word discovery.

\section{Limitations}
\label{sec:limitations}

The greatest limitation of our findings towards practical application is the need for aligned phonemic labels, which are not easily available in most tasks.
Future work may investigate ways to apply these insights using approximate or inferred labels, obviating the need for precise phonemic alignments.

Additionally, as described in Section \ref{sec:method}, we excluded frames for which the corresponding phoneme/diphone/triphone fully contained the corresponding word when estimating the regression from phonemic factor labels to model embeddings.
This was done in order to avoid inadvertently residualizing out words in addition to phonemic context, by restricting the input to the regression model to cases where the predictor could not be construed as constituting a word representation.
In some cases, this has an effect on the resulting probe accuracies.
As shown in Table \ref{tab:results-val}, the phoneme accuracy after phoneme residualization remains above chance level (relative frequency of the modal phoneme is 11.6\%), suggesting that this method fails to fully remove phoneme identity information.
However, if we estimate the regression without the exclusion policy just described, the resulting phoneme residuals succeed in removing almost all phoneme information (accuracy below 16\%) for layers 1--10 of HuBERT and wav2vec~2.0.
This means that including the frames corresponding to monophonemic words in the regression is essential to full linear removal of phoneme information.
One potential reason is that some phoneme is overrepresented as a monophonemic word versus as a part of other words, and the exclusion of monophonemic words thus disproportionately hurts the model estimation for that phoneme.

\section{Acknowledgments}
This work was supported by the Natural Sciences and Engineering Research Council of Canada (NSERC) RGPIN-2022-04431, the Data Sciences Institute, and the Linguistics Graduate Research Award at the University of Toronto, as well as resources provided by Compute Ontario, Calcul Québec, and the Digital Research Alliance of Canada.

\section{Generative AI use disclosure}
Generative AI was not used to produce this manuscript.

\bibliographystyle{IEEEtran}
\bibliography{custom}

\end{document}